\documentclass[runningheads]{llncs}
\usepackage{arxiv}

\usepackage{graphicx}
\usepackage[T1, T2A]{fontenc}
\usepackage[utf8]{inputenc}

\usepackage{CJKutf8}
\begin{document}
\title{Statistical and Neural Methods for Cross-lingual Entity Label Mapping in Knowledge Graphs}
%
%
\author{Gabriel Amaral\inst{1}\orcidID{0000-0002-4482-5376} \and
Mārcis Pinnis\inst{2}\orcidID{0000-0001-6832-5600} \and
Inguna Skadiņa\inst{2}\orcidID{0000-0003-4787-7099} \and
Odinaldo Rodrigues\inst{1}\orcidID{0000-0001-7823-1034} \and
Elena Simperl\inst{1}\orcidID{0000-0003-1722-947X}}
\authorrunning{G. Amaral et al.}
%
\institute{King's College London, London WC2R 2LS, UK \\
\email{\{gabriel.amaral,odinaldo.rodrigues,elena.simperl\}@kcl.ac.uk} \and
Tilde, LV-1004 Rīga, Latvia\\
\email{\{marcis.pinnis,inguna.skadina\}@tilde.lv}}
\maketitle              

\begin{abstract}
Knowledge bases such as Wikidata amass vast amounts of named entity information, such as multilingual labels, which can be extremely useful for various multilingual and cross-lingual applications. However, such labels are not guaranteed to match across languages from an information consistency standpoint, greatly compromising their usefulness for fields such as machine translation. In this work, we investigate the application of word and sentence alignment techniques coupled with a matching algorithm to align cross-lingual entity labels extracted from Wikidata in 10 languages. Our results indicate that mapping between Wikidata's main labels stands to be considerably improved (up to $20$ points in F1-score) by any of the employed methods. We show how methods relying on sentence embeddings outperform all others, even across different scripts. We believe the application of such techniques to measure the similarity of label pairs, coupled with a knowledge base rich in high-quality entity labels, to be an excellent asset to machine translation.

\keywords{Entity Label Mapping  \and Knowledge Representation \and Multilinguality \and Data Quality}
\end{abstract}
\section{Introduction}
Knowledge bases, such as Wikidata~\cite{vrandevcic2014wikidata} and DBpedia~\cite{auer2007dbpedia}, have amassed large amounts of multilingual information about various concepts. These include various named entities (e.g., persons, organisations, and locations) which can be useful for various language technologies, such as named entity recognition~\cite{chen2022ustc}, multilingual dictionaries~\cite{turki2017using}, and machine translation~\cite{merhav2018design,moussallem2019utilizing}.

Most multilingual data stored in these knowledge bases has been crowdsourced by non-professionals in linguistic aspects, let alone in aspects of multilinguality. This raises data quality concerns despite the existence of proper guidelines on creating appropriate labels,\footnote{\url{https://www.wikidata.org/wiki/Help:Label}} as these are not always followed by editors. 
Additionally, linguistic, regional, and cultural factors contribute to main labels assigned to an entity across languages not being fully correct or cross-lingually equivalent. For instance, the Wikidata entry for Donald Trump has the main label ``\textit{Donald Trump}'' in Lithuanian, however, the correct representation of the name in Lithuanian is ``\textit{Donaldas Trampas}''. The Wikidata entry for John F. Kennedy has the English label ``\textit{John F. Kennedy}'' and the Latvian label ``\textit{Džons Kenedijs}'' (without the initial of the middle name).

Besides main labels, Wikidata features also aliases, which are alternatives and variants that refer to that same entity. One entity's label in a certain language might correspond to an alias in another language. 
For instance, the American actor and politician Kane has the English label ``\textit{Kane}'' and the French label ``\textit{Glenn Jacobs}'', however, both labels can be also found in the list of alternative labels of the other respective language. The main labels and the aliases are not in any way cross-lingually mapped, which hinders automated use of the multilingual data in use cases that rely on high-quality cross-lingual dictionaries. 

Current neural machine translation (NMT) methods provide means for integration of term and named entity dictionaries in NMT systems thereby enforcing term and named entity translation consistency and correctness. E.g., the work by Bergmanis and Pinnis \cite{bergmanis2021facilitating} allows integrating terms and named entities in dictionary/canonical forms in NMT for morphologically rich languages. For these methods to work properly, it is crucial that translation equivalents represent adequate translations such that no information is lost or added with respect to the source language. 
If we only used Wikidata's main labels as a dictionary for machine translation, we would often produce wrong translations. For instance, the main label for the American entertainment company Miramax in English is ``\textit{Miramax}'' and in German -- ``\textit{Miramax Films}''. 
Translating the English sentence ``\textit{Miramax Films released a new movie}'' through a machine translation system that uses these main labels in its dictionary would yield ``\textit{Miramax Films Filme haben einen neuen Film veröffentlicht}'', with the word ``\textit{Films}'' translated twice.

Therefore, the focus of this work is on how to cross-lingually map Wikidata labels (both main labels and aliases) such that it is possible to acquire linguistically and semantically correct parallel named entity dictionaries from Wikidata. Our contributions are as follows:
\begin{itemize}
    \item We build and release a cross-lingual entity label mapping dataset based on Wikidata in order to aid research, ours and future, into improving entity label mapping.
    \item We apply and compare different cross-lingual word and sentence similarity metrics for the task of cross-lingual entity label mapping within Wikidata, demonstrating how sentence embedding techniques can greatly improve Wikidata's label mapping.
    \item We analyse the level of cross-lingual parallelism of main labels in Wikidata for 10 languages and show that cross-lingual data quality is a current issue in the knowledge base.
    \item We propose a method for cross-lingual mapping of labels that relies on similarity scores from cross-lingual word alignment methods and achieves a mapping accuracy of over $88\%$ on our Wikidata dataset.
\end{itemize}

The paper is structured as follows: Section~\ref{sec-rel-work} describes related work on data quality in Wikidata and cross-lingual mapping of entities, Section~\ref{sec-data-prep} described our benchmark dataset used in the experiments, Section~\ref{sec-mapping} describes our method for cross-lingual label mapping, Section~\ref{sec-res} describes and discusses the results achieved, and finally Section~\ref{sec-conclusions} concludes the paper.

\section{Related Work}
\label{sec-rel-work}

The quality and reliability of crowdsourced data have been discussed a lot in recent years, with different dimensions and metrics proposed and analysed~\cite{Daniel_2019,Adler_2011,Lewoniewski_2020}. However, few works analyse the quality of crowdsourced knowledge graph data, and only recent studies do so in a systematic way.

Most studies into measuring Wikidata's quality and identifying potential improvements 
ignore multilingual aspects~\cite{Shenoy_2021,piscopo2017provenance,mora2019systematic}. For instance, Skenoy et al.~\cite{Shenoy_2021} rely on language-agnostic editor behaviour and ontological properties to identify low-quality statements. Piscopo et al.~\cite{piscopo2017provenance} investigate the quality of information provenance for references only in English. 


Some recent studies do address the multilinguality of Wikidata. Kaffee et al.~\cite{Kaffee_2017} analyse language coverage in Wikidata, concluding that Wikidata knowledge is available in just a few languages, while many languages have almost no coverage. 
Amaral et al.~\cite{amaral2021assessing} investigate provenance quality across different languages with distinct coverage, finding that quality did not correlate with coverage and varied significantly between languages. As far as we know, ours is the first work to measure an intrinsically multilingual quality dimension of Wikidata from a standpoint of applicability in downstream language tasks.

The task of mapping entity labels in Wikidata is closely related to cross-lingual terminology and named entity extraction and mapping in comparable~\cite{cstefuanescu2012mining,daille2012building,dejean2002bilingual} or parallel corpora~\cite{lefever2009language}. Although these tasks are broader, involving the identification of words and phrases constituting terms and named entities in larger contexts, a crucial component in these tasks is the assessment of the cross-lingual parallelism of term and named entity phrases.
\c{S}tef\u{a}nescu~\cite{cstefuanescu2012mining} proposed a term similarity metric that combines probabilistic dictionary and cognate-based similarity scores.
While \c{S}tef\u{a}nescu
analysed terms on word level, Pinnis~\cite{Pinnis2013} proposed to align subwords between source and target terms and assess parallelism using a Levenshtein distance~\cite{Levenshtein1966} similarity metric. The sub-word-level nature of the method allows it to map compounds and complex multi-word terms.
Daille~\cite{daille2012building} assesses term co-occurrence statistics when calculating the parallelism of terms.
Aker et al.~\cite{aker2013extracting} train a binary classifier that predicts the parallelism of terms given a set of features including dictionary-based and cognate-based features.

Since in our work, we try to address mapping in a use case without contextual information, we compare the context-independent term mapping method by Pinnis
with more novel (state-of-the-art) text similarity assessment methods that rely on large pre-trained language models.

State-of-the-art methods usually rely on large, multilingual, pre-trained language models, such as BERT~\cite{Devlin2018}, XLM-R~\cite{conneau-etal-2020-unsupervised}, and ALBERT~\cite{Lan2020ALBERTAL}.
For example, SimAlign word alignments obtained from such models demonstrated better results for English-German than traditional statistical alignment methods~\cite{simalign}. 
Pre-trained multilingual BERT models are also used for cross-lingual alignment of multilingual knowledge graphs~\cite{DBLP:conf/emnlp/YangZSLLS19}.
Promising results have been demonstrated by sentence-level embedding methods, such as LaBSE~\cite{labse} and LASER~\cite{laser}.

The complexity of entity linking is recognized also by the \textit{Cross-lingual Challenge on Recognition, Normalization,
Classification, and Linking of Named Entities across Slavic languages}~\cite{piskorski-etal-2021-slav}. The task involved recognizing named entity mentions in Web documents, name normalization, and cross-lingual linking for six languages (Bulgarian, Czech, Polish, Russian, Slovene, Ukrainian). While the best model for this task in terms of F-score reached $85.7\%$ and performance for the named entity recognition task reached $90\%$ F-score, results for cross-lingual entity linking were not so promising, reaching only an F-score of $50.4\%$ for the best system~\cite{viksna-skadina-2021-multilingual} employing LaBSE sentence embeddings.

Finally, recent work by researchers from \textit{Google} proposes one dual encoder architecture model for linking 104 languages against 20 million Wikidata entities~\cite{DBLP:conf/emnlp/BothaSG20}. While the authors demonstrate the feasibility of the proposed approach, the quality and reliability of Wikidata are not discussed.

\section{Data Preparation}
\label{sec-data-prep}


To properly assess the effectiveness of entity label mapping methods, we need to construct a benchmark dataset from Wikidata.
We start by acquiring a full dump of the knowledge graph (as of November 2021). We identify the three main classes whose subclass trees will encompass our target entities: Person (\texttt{Q215627}), Organisation (\texttt{Q43229}), and Geographic Location (\texttt{Q2221906}). 
By following the paths defined by the graph's ``subclass of'' (\texttt{P279}) and ``instance of'' (\texttt{P31}) predicates, we identify and extract approximately 43K subclasses and 9.3M entities that are instances of Person, 27K subclasses and 3M instances of Organisation, and 29K subclasses and 10.3M instances of Place. In total, we extracted 21.6M distinct entities.

Our entity label dataset should not be too sparse, otherwise, our results would be unreasonably biased by dissimilar levels of label coverage between languages. Thus, we follow two approaches: keeping only languages with higher coverage of labels and aliases, as well as keeping only richly labelled entities.
For each language \textit{L}, we measure: its main label coverage, defined as the percentage of all entities that have a main label in \textit{L}, its alias presence, defined by the percentage of all entities that have at least one alias in \textit{L}, and the average quantity of aliases all entities have in \textit{L}. We rank all languages according to these metrics, calculate the average rank, and pick the top 10 languages to compose our dataset. They are: Swedish (\textit{SV}), German (\textit{DE}), Spanish (\textit{ES}), Russian (\textit{RU}), French (\textit{FR}), Italian(\textit{IT}), English (\textit{EN}), Portuguese (\textit{PT}), Chinese (\textit{ZH}), and Dutch (\textit{NL}).

We also filter entities based on their label coverage. To be kept, an entity must adhere to the following criteria: having a main label in at least 4 of the 10 selected languages and having at least 3 aliases in 3 of the 10 selected languages. These constraints are the highest values before label and alias coverage start to plateau. Out of 21.6M entities, only 33K ($0.16\%$) adhere to this criteria.

As Wikidata is a collaborative effort, labels or aliases may be put under the wrong language either by mistake or intentionally. Thus, final filtering is performed on the 33K extracted entities. We ascertain the languages of labels with fastText's~\cite{fastext1,fastext2} language detection models, which calculate the probabilities of a label belonging to each supported language. Labels that do not have an ambiguous language (e.g., acronyms, personal names, etc.) and have a very low probability of being of the language they are assigned to get dropped.

We finish by reorganising the dataset so that each entry consists of a unique cross-lingual pairing of labels for a given entity, including both main labels and aliases. Table~\ref{tab:sample} shows a small random sample of the dataset. Our benchmark dataset consists of 8.9M cross-lingual label pairings extracted from 33K entities in the 10 selected languages and is available online\footnote{\url{https://doi.org/10.6084/m9.figshare.19582798}}. The majority of entities extracted ($67\%$) are Persons, followed by Organisations ($22\%$) and Places ($10\%$). For every selected language, entities with main labels far outnumber those without, with 5 out of 10 having over $90\%$ coverage. The mean alias count is above $1$ for all languages (and above 2.5 for 5), and the alias coverage is around or above $50\%$ for all except \textit{ZH} ($37\%$). We perceive a moderate correlation ($0.57$) between the presence of \textit{RU} and \textit{ZH} labels, as well as between \textit{SV} and \textit{IT} ($0.47$).

\begin{table}[t!]
\caption{Small random sample from our benchmarking dataset. Each entry consists of a unique cross-lingual pair for a specific entity.}
\centering
\begin{tabular}{|l|l|l|l|l|}
\hline
\textbf{entity\_id} & \textbf{la\_1} & \textbf{lan\_2} & \textbf{label\_1} & \textbf{label\_2} \\ \hline
Q152265  & FR & ZH & Zaher Shah            & \begin{CJK*}{UTF8}{gbsn}穆罕默德·查希爾·沙阿\end{CJK*}       \\ \hline
Q1241726 & IT & PT & Rebecca Flanders      & Donna Ball        \\ \hline
Q1400802 & RU & EN & Иераполь Бамбика      & Manbug            \\ \hline
Q150652  & SV & DE & Vilhelm I av Tyskland & Kartätschenprinz  \\ \hline
Q275715  & ES & NL & Estadio de Montjuïc   & Olympisch Stadion \\ \hline
\end{tabular}
\label{tab:sample}
\end{table}

\section{Entity Label Mapping}
\label{sec-mapping}

We employ multiple methods to estimate the cross-lingual similarity of each entity label pair in our dataset and to solve the problem of entity label mapping. Then, we devise a greedy algorithm that transforms these scores into a set of high-similarity non-overlapping pairings of labels for each unique (\textit{entity id}, \textit{language 1}, \textit{language 2}) tuple. Finally, we measure the performance of each method and its variations in identifying these optimal pairings of labels in our Wikidata benchmark dataset and compare them. Please note that all label pre-processing is done by the methods themselves.

\subsection{Label Cross-Lingual Similarity Scoring methods}
MPAlligner~\cite{Pinnis2013} is a statistical cross-lingual terminology mapping tool that uses probabilistic dictionaries and Romanisation-based transliteration methods to identify reciprocal mappings between words and sub-word units of source and target terms. It scores each term pair by assessing the proportion of overlapping characters. Since MPAligner relies on the existence of large bilingual dictionaries or cognates that are shared between languages, its recall can be limited.

Simalign~\cite{simalign} is a word alignment method that does not require parallel data, relying instead on multilingual word embeddings created from monolingual data. It uses a multilingual BERT to retrieve sub-word embeddings and matches them across two sentences through a combination of cosine similarity and various matching strategies defined by the authors. We extract from Simalign the calculated similarity scores between sub-word units after being transformed by these matching strategies. Finally, we average out the pair's sub-word scores.

LASER~\cite{laser} and LaBSE~\cite{labse} are both sentence embedding methods. By embedding the entirety of each label, we acquire pairs of vectors to which we can directly apply similarity metrics. LASER embeds over 90 languages in a joint space by training an LSTM-based encoder-decoder neural network that is shared between all languages. LaBSE follows a similar approach but uses a multilingual BERT model fine-tuned on a translation ranking task. LaBSE establishes the current state-of-the-art results on multiple downstream multilingual tasks.

\subsection{Best Match Algorithm}
\label{sec-bestmatch}
We aim to find the best cross-lingual mapping between entity labels. Thus, we devise a greedy algorithm that, given a list of cross-lingual label pairings and their respective similarity scores for a specific entity, provides us with a non-overlapping set of pairings deemed as the best matches. We apply this algorithm to the scores produced by each scoring method, compiling distinct lists of best matches, and comparing each to a manually annotated ground truth.

First, the algorithm divides the dataset into groups of scored cross-lingual label pairs, each indexed by a unique (\textit{entity id}, \textit{language 1}, \textit{language 2}) tuple. On each group, it visits all pairs in descending order of similarity, declaring a pair as a best match only if no other pair in that same group containing either of this pair's labels was declared a best match before. This creates a one-to-one mapping between an entity's labels in a language $L1$ and language $L2$. If the entity does not have an equal amount of labels in both languages, some labels will remain without a match. This is expected and welcomed as not all labels have clear cross-lingual matches in Wikidata, and is better than perhaps forcing a match with dissimilar labels.

In addition to the aforementioned methods, we add two simple baselines: randomised and main label matching. The first declares best matches via a randomised walk through the dataset, rather than ordered by scores. The second declares all, and only, pairs of two main labels as best matches.

\subsection{Ground Truth and Method Comparison}
\label{sec-comparison}

The best match selection algorithm assigns pairs a binary class, i.e. best match or not. Thus, these mappings can be compared to ground truth and have their performances measured through standard classification metrics. Elaborating such a ground truth is not trivial, as we first need to define what truly constitutes a label's best match.

We define a best match as a cross-lingual pairing of labels for the same entity wherein one neither adds information to nor removes information from the other. By information, we mean any data about an entity's identity, attributes, or qualities. An example of best matches in our ground truth can be seen in Table~\ref{tab:best_matches_ground_truth}. This definition allows for potentially overlapping pairs, as Table~\ref{tab:best_matches_ground_truth} shows for entity \textit{Q311374}, in the case of minor variations such as spelling of transliterated sub-words that are commonly found on natural text. This means that none of the tested mapping methods can achieve the maximum classification accuracy, but it still allows us to compare their performance to each other as they still benefit from selecting one of the ground truth best matches.

\begin{table}[t!]
\caption{An example of pairings we consider best matches in our ground truth. For entity \textit{Q398}, the second and third pairs introduce information, namely the official acronym and the fact Bahrain is a kingdom. For entity \textit{Q311374}, pair 3 replaces the full first name with a generic nickname.}
\centering
\begin{tabular}{|l|l|l|l|l|l|}
\hline
\textbf{Entity} & \textbf{Lan1} & \textbf{Lan2} & \textbf{Label1} & \textbf{Label2} & \textbf{Best} \\ \hline
Q398   & SV & EN & Bahrain              & Bahrain                                                                  & \textbf{Yes} \\ \hline
Q398   & SV & EN & Bahrain              & BAH                                                                      & No           \\ \hline
Q398   & SV & EN & Bahrain              & Kingdom of Bahrain                                                       & No           \\ \hline
Q398   & SV & EN & Konungariket Bahrain & Kingdom of Bahrain                                                       & \textbf{Yes} \\ \hline
Q311374 & SV & FR & Aleksandr Ovetjkin    & Aleksandr Ovetchkine                                                        & \textbf{Yes} \\ \hline
Q311374 & SV & FR & Aleksandr Ovetjkin    & Alexander Ovechkin & \textbf{Yes}           \\ \hline
Q311374 & SV & FR & Aleksandr Ovetjkin    & Alex Ovechkin                                                        & No \\ \hline
\end{tabular}
\label{tab:best_matches_ground_truth}
\end{table}

Our ground truth consists of a manually annotated representative sample from our benchmark dataset ($95\%$ confidence interval, $5\%$ margin of error) rather than its totality due to annotation costs. 
This is obtained through a stratified sampling after a few processing steps to account for underlying aspects of the label distribution which might hinder comparison between methods. That is:

\begin{enumerate}
    \item We remove from the benchmark dataset all entities with an outlier amount of labels pairs so that they do not bias the comparison. E.g., the entity \textit{``Pope Adeodato I''}, has over 42 \textit{PT}-\textit{ZH} label pairs, over six times the average.
    \item We remove all entities containing only one label pair for any language combination, as they are trivially solvable and do not contribute to the comparison.
    \item We randomly reduce the number of entities with an identical pair (e.g. ``Bahrain'' and ``Bahrain'' in Table~\ref{tab:best_matches_ground_truth}) by a factor of $50\%$, as they represent a much smaller challenge than entities with no identical pairs, yet compose half of all entities. This lets our comparison focus more on harder cases.
\end{enumerate}

Then, we perform a stratified sampling so that languages are equally represented and carry similar weight. We manually annotate this sample according to our definition of best match to compose our ground truth.

\section{Evaluation and Discussion}
\label{sec-res}


With our ground truth, we can measure the performance of each of the cross-lingual label similarity estimation methods and their variations. From the four methods described in Section~\ref{sec-mapping}, we devise and test nine variations, except for MPAlligner, which is used as-is. For Simalign, we use each of its matching methods (Argmax, Match, and Itermax) to filter out scores of non-matching sub-word pairings and average those remaining. We also extract the sub-word embeddings used by the method and apply cosine similarity directly to them, extracting the mean. As for both LASER and LaBSE, we calculate cosine similarity and inverse Euclidean distance between pairs of label embeddings.

\begin{figure}[t!]
\begin{minipage}[c]{0.49\linewidth}
\includegraphics[width=\linewidth]{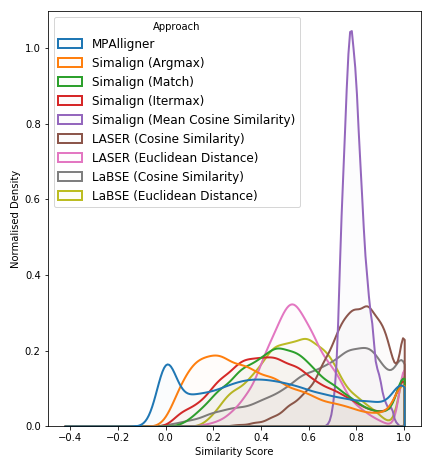}
\caption{Normalised probability density of the similarity scores calculated by each approach variation.}
\label{fig:score_dist}
\end{minipage}
\hfill
\begin{minipage}[c]{0.49\linewidth}
\includegraphics[width=\linewidth]{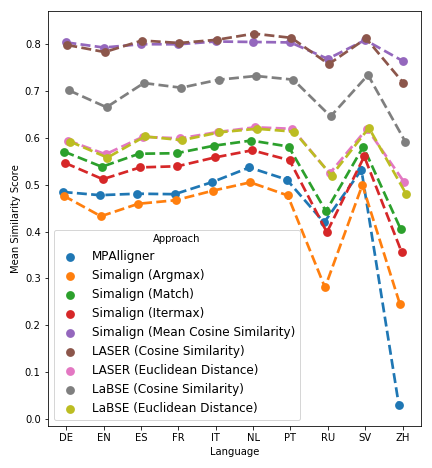}
\caption{The mean similarity scores per language as calculated by each approach variation.}
  \label{fig:mean_score_per_lan}
\end{minipage}%
\end{figure}

Figure~\ref{fig:score_dist} shows the density distributions of similarity scores from each method. We can see how most methods have a bell-shaped curve between values 0 to 1, with a spike near 1, due to many pairs being nearly or truly identical. MPAlligner has a spike near 0; as it depends on constructed dictionaries, the lack of explicit equivalences means it will default to low values of similarity. The mean cosine similarity applied at Simalign embeddings has a narrow and tall curve centred in a high score, indicating these scores are either not very informative, or their domain is naturally restricted. All other methods follow expected distributions.

Figure~\ref{fig:mean_score_per_lan} shows the mean similarity scores per language. All approaches seem to calculate similar scores regardless of language, except for the only two languages not using the Latin script: \textit{RU} and \textit{ZH}. This drop might be attributed to two factors. One is the natural lessened similarity between labels using different scripts, the other is the methods' inability to perceive cross-script similarities. This drop is bigger with the dictionary-based method (MPAlligner), lesser with sub-word embeddings, and minimal with sentence embeddings.


In Table~\ref{tab:results}, we show the accuracy obtained by all methods for all the data in the ground truth annotation, as well as broken down by language, including both baselines. Looking at the main-label baseline, we see how current Wikidata main labels can be significantly improved in terms of cross-lingual matching. For most languages tested, the gap in accuracy between depending on main labels and automated methods is very high, e.g., for \textit{IT}, \textit{FR}, \textit{ES}, etc.

We can see sentence-embedding models performing far better than other methods. MPAlligner generally outperforms sub-word embedding methods, which we find surprising, and indicates that approaching this task by looking at labels in their entirety rather than broken into units is beneficial. All methods are generally better than the baselines, except in the case of non-Latin script languages \textit{RU} and \textit{ZH}. For \textit{RU}, most Simalign-based methods under-perform the main-label baseline, whereas for \textit{ZH} that is all Simalign-based methods, some under-performing even the randomised baseline. Still, for \textit{ZH}, LaBSE with cosine similarity is the only approach significantly outperforming the main-label baseline. These results point to LaBSE as the best-performing method for cross-lingual entity label mapping.



\begin{table}[t!]
\caption{Classification performance of different approaches as measured by accuracy across all languages and individually. \textit{MPA} = MPAlligner; \textit{SIM\_A} = Simalign (Argmax); \textit{SIM\_M} = Simalign (Match); \textit{SIM\_I} = Simalign (Itermax); \textit{SIM\_C} = Simalign (Cosine); \textit{LS\_C} = LASER (Cosine); \textit{LS\_E} = LASER (Euclidean); \textit{LB\_C} = LaBSE (Cosine); \textit{RAN} = Randomised Baseline; \textit{ML} = Main Labels Baseline. LaBSE with Euclidean Similarity performs identically to LaBSE with Cosine Similarity.}
\centering
\begin{tabular}{|l|l|l|l|l|l|l|l|l|l|l|l|}
\hline
\textbf{} &
  \textbf{ALL} &
  \textbf{DE} &
  \textbf{EN} &
  \textbf{ES} &
  \textbf{FR} &
  \textbf{IT} &
  \textbf{NL} &
  \textbf{PT} &
  \textbf{RU} &
  \textbf{SV} &
  \textbf{ZH} \\ \hline
\textbf{MPA}    & 0.832 & 0.846 & 0.840 & 0.830 & 0.816 & 0.867 & 0.880 & 0.833 & 0.865 & 0.813          & 0.679 \\ \hline
\textbf{SIM\_A} & 0.827 & 0.826 & 0.844 & 0.798 & 0.844 & 0.857 & 0.880 & 0.860 & 0.865 & 0.801          & 0.649 \\ \hline
\textbf{SIM\_M} & 0.808 & 0.851 & 0.831 & 0.807 & 0.844 & 0.814 & 0.817 & 0.827 & 0.806 & 0.832          & 0.582 \\ \hline
\textbf{SIM\_I} & 0.810 & 0.841 & 0.844 & 0.798 & 0.844 & 0.809 & 0.805 & 0.838 & 0.822 & 0.819          & 0.619 \\ \hline
\textbf{SIM\_C} & 0.807 & 0.831 & 0.836 & 0.816 & 0.816 & 0.835 & 0.828 & 0.822 & 0.822 & 0.795          & 0.597 \\ \hline
\textbf{LS\_C} &
  0.871 &
  0.896 &
  \textbf{0.900} &
  \textbf{0.887} &
  0.877 &
  0.873 &
  \textbf{0.902} &
  0.854 &
  \textbf{0.908} &
  0.863 &
  0.694 \\ \hline
\textbf{LS\_E}  & 0.867 & 0.891 & 0.896 & 0.873 & 0.883 & 0.873 & 0.880 & 0.870 & 0.892 & \textbf{0.869} & 0.686 \\ \hline
\textbf{LB\_C} &
  \textbf{0.882} &
  \textbf{0.905} &
  0.879 &
  \textbf{0.887} &
  \textbf{0.894} &
  \textbf{0.925} &
  0.880 &
  \textbf{0.887} &
  0.897 &
  0.863 &
  \textbf{0.768} \\ \hline
\textbf{RAN}    & 0.656 & 0.633 & 0.698 & 0.657 & 0.661 & 0.656 & 0.657 & 0.591 & 0.758 & 0.621          & 0.604 \\ \hline
\textbf{ML}     & 0.776 & 0.787 & 0.784 & 0.765 & 0.755 & 0.735 & 0.788 & 0.779 & 0.844 & 0.819          & 0.679 \\ \hline
\end{tabular}
\label{tab:results}
\end{table}

\section{Conclusion}
\label{sec-conclusions}

Cross-lingual mapping of entity labels such that information is properly preserved is an important challenge to be solved if we wish downstream tasks depending on such entities to improve. Resources such as Wikidata can greatly help, provided their labels have a higher quality of cross-lingual alignment. Through our contributions, we not only showcase the importance of such resources but suggest methods to improve such quality.

In this paper, we have presented the case of Wikidata by extracting and structuring a benchmark cross-lingual entity label mapping dataset from thousands of its entities. We have showcased a comparison between the performances of various text similarity estimation methods when applied to the task of cross-lingual entity label mapping. This comparison consists of adapting the various text similarity methods so that similarity scores are extracted for label pairs; devising an algorithm to select best matches based on similarity scores; developing a balanced and expressive ground truth dataset for the proper comparison of classification metrics.

We verified that Wikidata's main labels overall fail to match cross-lingual label pairings better than any of the text similarity estimation methods that were tested. We also ascertained that methods applied to labels as a whole tend to outperform those focused on their word and sub-word units. Furthermore, techniques based on sentence embeddings learned in a shared multilingual space have not only considerably outperformed other methods in same-script pairings, but also between distinct scripts. Finally, we have seen how many current and sophisticated word alignment techniques under-perform simplistic baselines at this task in specific languages. Our best match algorithm is based on comparisons between scores only, which is why we removed from our ground truth all entities with a single pairing on the basis of being trivially resolvable, even if that pairing is wrong. Using a threshold-sensitive approach would better treat these cases and is an interesting direction for future work.

\subsubsection*{Acknowledgements} This research received funding from the European Union’s Horizon 2020 research and innovation programme under the Marie Skłodowska-Curie grant agreement no. 812997. The research has been supported by the European Regional Development Fund within the research project ``AI Assistant for Multilingual Meeting Management'' No. 1.1.1.1/19/A/082.

%
%
\bibliographystyle{splncs04}
\bibliography{references}

\begin{thebibliography}{10}
\providecommand{\url}[1]{\texttt{#1}}
\providecommand{\urlprefix}{URL }
\providecommand{\doi}[1]{https://doi.org/#1}

\bibitem{Adler_2011}
Adler, B.T., de~Alfaro, L., Mola{-}Velasco, S.M., Rosso, P., West, A.G.:
  Wikipedia {V}andalism {D}etection: {C}ombining {N}atural {L}anguage,
  {M}etadata, and {R}eputation {F}eatures. In: Gelbukh, A.F. (ed.)
  Computational Linguistics and Intelligent Text Processing (CICLing 2011).
  Proceedings, Part {II}. Lecture Notes in Computer Science, vol.~6609, pp.
  277--288. Springer (2011)

\bibitem{aker2013extracting}
Aker, A., Paramita, M.L., Gaizauskas, R.: Extracting {B}ilingual
  {T}erminologies from {C}omparable {C}orpora. In: Proceedings of the 51st
  Annual Meeting of the Association for Computational Linguistics (Volume 1:
  Long Papers). pp. 402--411 (2013)

\bibitem{amaral2021assessing}
Amaral, G., Piscopo, A., Kaffee, L.a., Rodrigues, O., Simperl, E.: Assessing
  the {Q}uality of {S}ources in {W}ikidata {A}cross {L}anguages: {A} {H}ybrid
  {A}pproach. Journal of Data and Information Quality  \textbf{13}(4) (oct
  2021)

\bibitem{laser}
Artetxe, M., Schwenk, H.: Massively {M}ultilingual {S}entence {E}mbeddings for
  {Z}ero-{S}hot {C}ross-{L}ingual {T}ransfer and {B}eyond. Transactions of the
  Association for Computational Linguistics  \textbf{7},  597--610 (2019)

\bibitem{auer2007dbpedia}
Auer, S., Bizer, C., Kobilarov, G., Lehmann, J., Cyganiak, R., Ives, Z.:
  Dbpedia: a {N}ucleus for a {W}eb of {O}pen {D}ata. In: The Semantic Web, pp.
  722--735. Springer (2007)

\bibitem{bergmanis2021facilitating}
Bergmanis, T., Pinnis, M.: Facilitating terminology translation with target
  lemma annotations. In: Proceedings of the 16th Conference of the European
  Chapter of the Association for Computational Linguistics: Main Volume. pp.
  3105--3111 (2021)

\bibitem{DBLP:conf/emnlp/BothaSG20}
Botha, J.A., Shan, Z., Gillick, D.: Entity {L}inking in 100 {L}anguages. In:
  Webber, B., Cohn, T., He, Y., Liu, Y. (eds.) Proceedings of the 2020
  Conference on Empirical Methods in Natural Language Processing, {EMNLP} 2020.
  pp. 7833--7845 (2020)

\bibitem{chen2022ustc}
Chen, B., Ma, J.Y., Qi, J., Guo, W., Ling, Z.H., Liu, Q.: {USTC-NELSLIP} at
  {S}em{E}val-2022 {T}ask 11: {G}azetteer-{A}dapted {I}ntegration {N}etwork for
  {M}ultilingual {C}omplex {N}amed {E}ntity {R}ecognition. arXiv:2203.03216
  (2022)

\bibitem{conneau-etal-2020-unsupervised}
Conneau, A., Khandelwal, K., Goyal, N., Chaudhary, V., Wenzek, G., Guzm{\'a}n,
  F., Grave, E., Ott, M., Zettlemoyer, L., Stoyanov, V.: Unsupervised
  {C}ross-lingual {R}epresentation {L}earning at {S}cale. In: Proceedings of
  the 58th Annual Meeting of the Association for Computational Linguistics. pp.
  8440--8451 (2020)

\bibitem{daille2012building}
Daille, B.: Building {B}ilingual {T}erminologies from {C}omparable {C}orpora:
  {T}he {TTC} {T}erm{S}uite. In: 5th Workshop on Building and Using Comparable
  Corpora (2012)

\bibitem{Daniel_2019}
Daniel, F., Kucherbaev, P., Cappiello, C., Benatallah, B., Allahbakhsh, M.:
  Quality {C}ontrol in {C}rowdsourcing: {A} {S}urvey of {Q}uality {A}ttributes,
  {A}ssessment {T}echniques, and {A}ssurance {A}ctions. ACM Comput. Surv.
  \textbf{51}(1) (jan 2018)

\bibitem{dejean2002bilingual}
D{\'e}jean, H., Gaussier, {\'E}., Sadat, F.: Bilingual {T}erminology
  {E}xtraction: {A}n {A}pproach {B}ased on a {M}ultilingual {T}hesaurus
  {A}pplicable to {C}omparable {C}orpora. In: Proceedings of the 19th
  International Conference on Computational Linguistics COLING. pp. 218--224
  (2002)

\bibitem{Devlin2018}
Devlin, J., Chang, M.W., Lee, K., Toutanova, K.: {BERT: Pre-training of Deep
  Bidirectional Transformers for Language Understanding}. In: Proceedings of
  the 2019 Conference of the North American Chapter of the Association for
  Computational Linguistics: Human Language Technologies, Volume 1 (Long and
  Short Papers). pp. 4171--4186 (2019)

\bibitem{labse}
Feng, F., Yang, Y., Cer, D., Arivazhagan, N., Wang, W.: Language-agnostic
  {BERT} {S}entence {E}mbedding (2020)

\bibitem{simalign}
Jalili~Sabet, M., Dufter, P., Yvon, F., Sch{\"u}tze, H.: {S}im{A}lign: High
  quality word alignments without parallel training data using static and
  contextualized embeddings. In: Findings of the Association for Computational
  Linguistics: EMNLP 2020. pp. 1627--1643. Association for Computational
  Linguistics, Online (Nov 2020)

\bibitem{fastext2}
Joulin, A., Grave, E., Bojanowski, P., Douze, M., J{\'e}gou, H., Mikolov, T.:
  Fasttext.zip: Compressing text classification models. ArXiv
  \textbf{abs/1612.03651} (2016)

\bibitem{fastext1}
Joulin, A., Grave, E., Bojanowski, P., Mikolov, T.: Bag of tricks for efficient
  text classification. In: Proceedings of the 15th Conference of the {E}uropean
  Chapter of the Association for Computational Linguistics: Volume 2, Short
  Papers. pp. 427--431. Association for Computational Linguistics, Valencia,
  Spain (Apr 2017)

\bibitem{Kaffee_2017}
Kaffee, L.A., Piscopo, A., Vougiouklis, P., Simperl, E., Carr, L., Pintscher,
  L.: A {G}limpse into {B}abel: {A}n {A}nalysis of {M}ultilinguality in
  {W}ikidata. In: Proceedings of the 13th International Symposium on Open
  Collaboration. OpenSym '17 (2017)

\bibitem{Lan2020ALBERTAL}
Lan, Z., Chen, M., Goodman, S., Gimpel, K., Sharma, P., Soricut, R.: {ALBERT:}
  {A} lite {BERT} for self-supervised learning of language representations. In:
  8th International Conference on Learning Representations, {ICLR} 2020, Addis
  Ababa, Ethiopia, April 26-30, 2020 (2020)

\bibitem{lefever2009language}
Lefever, E., Macken, L., Hoste, V.: Language-independent {B}ilingual
  {T}erminology {E}xtraction from a {M}ultilingual {P}arallel {C}orpus. In:
  Proceedings of the 12th Conference of the European Chapter of the ACL (EACL
  2009). pp. 496--504 (2009)

\bibitem{Levenshtein1966}
Levenshtein, V.I.: {Binary Codes Capable of Correcting Deletions, Insertions,
  and Reversals}. Soviet Physics Doklady  \textbf{10}(8),  707----710 (1966)

\bibitem{Lewoniewski_2020}
Lewoniewski, W., W\k{e}cel, K., Abramowicz, W.: Modeling {P}opularity and
  {R}eliability of {S}ources in {M}ultilingual {W}ikipedia. Information
  \textbf{11}(5) (2020)

\bibitem{merhav2018design}
Merhav, Y., Ash, S.: Design {C}hallenges in {N}amed {E}ntity {T}ransliteration.
  In: Proceedings of the 27th International Conference on Computational
  Linguistics. pp. 630--640 (2018)

\bibitem{mora2019systematic}
Mora-Cantallops, M., S{\'a}nchez-Alonso, S., Garc{\'\i}a-Barriocanal, E.: A
  {S}ystematic {L}iterature {R}eview on {W}ikidata. Data Technologies and
  Applications  (2019)

\bibitem{moussallem2019utilizing}
Moussallem, D., Ngonga~Ngomo, A.C., Buitelaar, P., Arcan, M.: Utilizing
  {K}nowledge {G}raphs for {N}eural {M}achine {T}ranslation {A}ugmentation. In:
  Proceedings of the 10th International Conference on Knowledge Capture. pp.
  139--146 (2019)

\bibitem{Pinnis2013}
Pinnis, M.: {Context Independent Term Mapper for European Languages}. In:
  Proceedings of Recent Advances in Natural Language Processing (RANLP 2013).
  pp. 562--570 (2013)

\bibitem{piscopo2017provenance}
Piscopo, A., Kaffee, L.A., Phethean, C., Simperl, E.: Provenance {I}nformation
  in a {C}ollaborative {K}nowledge {G}raph: {A}n {E}valuation of {W}ikidata
  {E}xternal {R}eferences. In: d'Amato, C., Fernandez, M., Tamma, V., Lecue,
  F., Cudr{\'e}-Mauroux, P., Sequeda, J., Lange, C., Heflin, J. (eds.) The
  Semantic Web -- ISWC 2017. pp. 542--558. Springer International Publishing,
  Cham (2017)

\bibitem{piskorski-etal-2021-slav}
Piskorski, J., Babych, B., Kancheva, Z., Kanishcheva, O., Lebedeva, M.,
  Marci{\'n}czuk, M., Nakov, P., Osenova, P., Pivovarova, L., Pollak, S.,
  P{\v{r}}ib{\'a}{\v{n}}, P., Radev, I., Robnik-Sikonja, M., Starko, V.,
  Steinberger, J., Yangarber, R.: Slav-{NER}: {T}he 3rd {C}ross-lingual
  {C}hallenge on {R}ecognition, {N}ormalization, {C}lassification, and
  {L}inking of {N}amed {E}ntities across {S}lavic {L}anguages. In: Proceedings
  of the 8th Workshop on Balto-Slavic Natural Language Processing. pp. 122--133
  (2021)

\bibitem{Shenoy_2021}
Shenoy, K., Ilievski, F., Garijo, D., Schwabe, D., Szekely, P.A.: A {S}tudy of
  the {Q}uality of {W}ikidata. CoRR  \textbf{abs/2107.00156} (2021)

\bibitem{cstefuanescu2012mining}
{\c{S}}tef{\u{a}}nescu, D.: Mining for {T}erm {T}ranslations in {C}omparable
  {C}orpora. In: The 5th Workshop on Building and Using Comparable Corpora. pp.
  98--103 (2012)

\bibitem{turki2017using}
Turki, H., Vrandecic, D., Hamdi, H., Adel, I.: Using {W}ikidata as a
  {M}ulti-lingual {M}ulti-dialectal {D}ictionary for {A}rabic {D}ialects. In:
  2017 IEEE/ACS 14th International Conference on Computer Systems and
  Applications (AICCSA). pp. 437--442 (2017)

\bibitem{viksna-skadina-2021-multilingual}
V{\=\i}ksna, R., Skadina, I.: Multilingual {S}lavic {N}amed {E}ntity
  {R}ecognition. In: Proceedings of the 8th Workshop on Balto-Slavic Natural
  Language Processing. pp. 93--97 (2021)

\bibitem{vrandevcic2014wikidata}
Vrande{\v{c}}i{\'c}, D., Kr{\"o}tzsch, M.: Wikidata: {A} {F}ree {C}ollaborative
  {K}nowledgebase. Communications of the ACM  \textbf{57}(10),  78--85 (2014)

\bibitem{DBLP:conf/emnlp/YangZSLLS19}
Yang, H., Zou, Y., Shi, P., Lu, W., Lin, J., Sun, X.: Aligning
  {C}ross-{L}ingual {E}ntities with {M}ulti-{A}spect {I}nformation. In: Inui,
  K., Jiang, J., Ng, V., Wan, X. (eds.) Proceedings of the 2019 Conference on
  Empirical Methods in Natural Language Processing and the 9th International
  Joint Conference on Natural Language Processing, {EMNLP-IJCNLP} 2019. pp.
  4430--4440 (2019)

\end{thebibliography}
%
%
%
%
%


\end{document}